\documentclass[conference]{IEEEtran}
\IEEEoverridecommandlockouts
\usepackage{cite}
\usepackage{amsmath,amssymb,amsfonts}
\usepackage{algorithmic}
\usepackage{graphicx}
\usepackage{textcomp}
\usepackage{xcolor}
\usepackage{soul}
\usepackage{tcolorbox}
\usepackage[utf8]{inputenc}
\usepackage[T1]{fontenc}
\usepackage{algorithm}
\usepackage{algorithmic}
\usepackage{amsmath}
\usepackage{amsthm}
\usepackage{booktabs}
\usepackage[switch]{lineno}
\usepackage{microtype}
\usepackage{nicefrac}
\usepackage{xcolor}
\usepackage{sidecap}
\usepackage{stfloats}
\usepackage{wrapfig}
\usepackage{threeparttable}
\usepackage{color}
\usepackage{mathtools}
\usepackage{transparent}
\usepackage{titlesec} 
\usepackage{makecell}
\usepackage{url}
\usepackage{float}
\usepackage{subcaption}
\usepackage{graphicx}
\usepackage{multirow}
\usepackage{xspace}
\usepackage{enumitem}
\usepackage[font=small]{caption}
\usepackage{diagbox}
\newtheorem{definition}{Definition}
\newtheorem{remark}{Remark}

\newtheorem{theorem}{Theorem}
\newtheorem{assumption}{Assumption}
\newtheorem{corollary}{Corollary}
\makeatletter
\newcommand{\linebreakand}{%
  \end{@IEEEauthorhalign}
  \hfill\mbox{}\par
  \mbox{}\hfill\begin{@IEEEauthorhalign}
}
\makeatother

\usepackage[textsize=tiny]{todonotes}

\usepackage{hyperref}
\usepackage{graphicx}
\usepackage{booktabs}
\usepackage[accsupp]{axessibility}  % Improves PDF readability for those with disabilities.

\def\BibTeX{{\rm B\kern-.05em{\sc i\kern-.025em b}\kern-.08em
    T\kern-.1667em\lower.7ex\hbox{E}\kern-.125emX}}
\begin{document}

\title{FedMABA: Towards Fair Federated Learning through Multi-Armed Bandits Allocation}

\author{\IEEEauthorblockN{1\textsuperscript{st} Zhichao Wang}
\IEEEauthorblockA{\textit{School of Science and Engineering} \\
\textit{The Chinese University of }\\
\textit{Hong Kong (Shenzhen)}\\
Shenzhen, China \\
zhichaowang@link.cuhk.edu.cn}
~\\
\and
\IEEEauthorblockN{2\textsuperscript{nd} Lin Wang}
\IEEEauthorblockA{\textit{School of Science and Engineering} \\
\textit{The Chinese University of }\\
\textit{Hong Kong (Shenzhen)}\\
Shenzhen, China \\
linwang1@link.cuhk.edu.cn}
~\\
\and
\IEEEauthorblockN{3\textsuperscript{rd} Yongxin Guo}
\IEEEauthorblockA{\textit{School of Science and Engineering} \\
\textit{The Chinese University of }\\
\textit{Hong Kong (Shenzhen)}\\
Shenzhen, China \\
yongxinguo@link.cuhk.edu.cn}
\\
\linebreakand
\IEEEauthorblockN{4\textsuperscript{th} Ying-Jun Angela Zhang}
\IEEEauthorblockA{\textit{Department of Information Engineering} \\
\textit{The Chinese University of Hong Kong}\\
Hong Kong, China \\
yjzhang@ie.cuhk.edu.hk}

\and
\IEEEauthorblockN{5\textsuperscript{th} Xiaoying Tang*}
\IEEEauthorblockA{\textit{School of Science and Engineering} \\
\textit{The Chinese University of Hong Kong (Shenzhen)}\\
Shenzhen, China \\
tangxiaoying@cuhk.edu.cn}
*Corresponding author
}

% \author{
% \IEEEauthorblockN{Chen Yang,
% Jie Liu,
% Xi Leng,
% Xiaoying Tang}\\

% \IEEEauthorblockA{\quad Email: \{chenyang2,jieliu3,xileng\}@link.cuhk.edu.cn,tangxiaoying@cuhk.edu.cn}
% \thanks{Chen Yang and  Jie Liu contributed equally to this work.\\
% \indent Xiaoying Tang is the corresponding author (tangxiaoying@cuhk.edu.cn).\\
% \indent 
% Chen Yang, Jie Liu, and Xi Leng and Xiaoying Tang are with the School of Science and Engineering, The Chinese University of Hong Kong, Shenzhen, Guangdong, 518172, P.R. China, and the Shenzhen Institute of Artificial Intelligence and Robotics for Society, The Chinese University of Hong Kong, Shenzhen, Guangdong, 518172, P.R. China, and the Shenzhen Key Laboratory of Crowd Intelligence Empowered Low-Carbon Energy Network, The Chinese University of Hong Kong, Shenzhen, Guangdong, 518172, P.R. China. \\
% \indent 
% This work is supported in part by the funding from Shenzhen Institute of Artificial Intelligence and Robotics for Society, in part by the Shenzhen Key Lab of Crowd Intelligence Empowered Low-Carbon Energy Network (Grant No. ZDSYS20220606100601002), in part by Shenzhen Stability Science Program 2023, and in part by the Guangdong Provincial Key Laboratory of Future Networks of Intelligence (Grant No. 2022B1212010001).}
% }

\maketitle

\begin{abstract}
% The increasing concern for data privacy has driven the rapid development of federated learning (FL), a privacy-preserving collaborative paradigm. However, the statistical heterogeneity among clients in FL leads to inconsistent performance of the server model across various clients, heightening the challenge of fairness. In this paper, we reconsider the inconsistency in client performance distribution and introduce the concept of adversarial multi-armed bandit to solve the proposed optimization objective with explicit constraints on performance disparities. Practically, we introduce a novel multi-armed bandit-based allocation FL algorithm (FedMABA) to mitigate performance unfairness among diverse clients with different data distributions, and even different tasks. Extensive experiments in FL, including federated multi-task learning, demonstrate the exceptional performance of FedMABA in enhancing fairness.

The increasing concern for data privacy has driven the rapid development of federated learning (FL), a privacy-preserving collaborative paradigm. However, the statistical heterogeneity among clients in FL results in inconsistent performance of the server model across various clients. Server model may show favoritism towards certain clients while performing poorly for others, heightening the challenge of fairness. In this paper, we reconsider the inconsistency in client performance distribution and introduce the concept of adversarial multi-armed bandit to optimize the proposed objective with explicit constraints on performance disparities. Practically, we propose a novel multi-armed bandit-based allocation FL algorithm (FedMABA) to mitigate performance unfairness among diverse clients with different data distributions. Extensive experiments, in different Non-I.I.D. scenarios, demonstrate the exceptional performance of FedMABA in enhancing fairness.
\end{abstract}


\renewcommand{\thefootnote}{\relax} % 重置脚注编号
\footnotetext{Zhichao Wang, Lin Wang, Yongxin Guo and Xiaoying Tang are with the School of Science and Engineering, The Chinese University of Hong Kong, Shenzhen, Guangdong, 518172, P.R. China, and the Shenzhen Institute of Artificial Intelligence and Robotics for Society, The Chinese University of Hong Kong, Shenzhen, Guangdong, 518172, P.R. China, and the Shenzhen Key Laboratory of Crowd Intelligence Empowered Low-Carbon Energy Network, School of Science and Engineering, The Chinese University of Hong Kong, Shenzhen, Guangdong, 518172, P.R. China (e-mail: zhichaowang@link.cuhk.edu.cn; linwang1@link.cuhk.edu.cn; yongxinguo@link.cuhk.edu.cn; tangxiaoying@cuhk.edu.cn). Ying-Jun Angela Zhang is with the Department of Information Engineering, The Chinese University of Hong Kong, Shatin, Hong Kong (e-mail: yjzhang@ie.cuhk.edu.hk).} 

\section{Introduction}
\label{sec:intro}

Federated Learning (FL) allows edge devices to collaboratively build a model by sharing updated parameters with a server, ensuring data privacy without exposing raw data \cite{mcmahan2017communication}. However, data heterogeneity among clients poses challenges, leading to performance degradation, inefficient convergence, and unfairness \cite{guo2023fedbr, wang2023delta, guo2024fedrc, wang2023fedeba+}. Poor aggregation strategies can further harm individual contributions and server model performance, affecting client motivation \cite{wu2023incentive}. Addressing these challenges is key to FL's fairness and development. While fairness in FL has seen progress \cite{mohri2019agnostic, hu2020fedmgda+, li2019fair, li2020federated, wang2021federated}, directly minimizing performance variance among clients remains an NP-hard problem, probably resulting in suboptimal global model performance.

% However, most of the existing approaches promote fairness by only considering the performance of clients participating in a certain training round. Consequently, we pose the question: 

% \textit{Is it possible to design a method that adaptively iterates towards a fair model update, accommodating partial client participation, while considering for all clients?}

% There are encouraging results on fairness in FL, such as specific objectives optimizing based algorithms \cite{mohri2019agnostic, hu2020fedmgda+}, global objective balancing optimization algotithms \cite{li2019fair, li2020federated}, and gradients correction based aggregation\cite{wang2021federated}. However, ensuring fairness in FL while maintaining the performance of the global model still remains an open challenge. Existing approaches either cannot significantly improve fairness or have a notable impact on the overall model performance. we thus pose a question: \textit{\textbf{Can we design a method aimed at significantly improving fairness among clients while not affecting the overall performance of the global model?}}

In this work, we design a novel FL algorithm: \textbf{Fed}erated Learning through \textbf{M}ulti-\textbf{A}rmed \textbf{B}andits \textbf{A}llocaiton named FedMABA. Specifically, we:

% Specifically, we:
% \begin{enumerate}[leftmargin=12pt,nosep]
% \setlength{\itemsep}{3.5pt}
%     \item First introduce explicit constraints on client performance distribution consistency in FL, transform the NP-hard optimization objective into a convex upper surrogate, and optimize it via the concept of adversarial multi-armed bandits.
%     \item Design a novel update strategy, combining the fair weights computed by loss variance constraints with the average ones, to ensure the performance of the server model.
%     \item Theoretically prove the generalization performance of server model is upper bounded by the variance of clients' performance, which demonstrate that constraining the performance distribution among clients helps improve the generalization performance of the server model. We offer the convergence analysis for FedMABA, and show it converges to the stationary point at a rate of $\mathcal{O}(\frac{1}{\sqrt{NKT}} + \frac1T)$.
%     % .
%     \item Conduct extensive experiments on Fashion-MNIST, CIFAR-10, and CIFAR-100 within FL, demonstrate that FedMABA maintains competitive server model performance and outperforms baselines in terms of fairness. In addition, we  do ablation experiments to assess the impact of hyper-parameters on FedMABA, and showcase the stability of FedMABA with different hyper-parameters and scenarios.
% \end{enumerate}

\begin{enumerate}[leftmargin=12pt,nosep]
\setlength{\itemsep}{3.5pt}
    \item First introduce explicit constraints on client performance distribution in FL, transform the NP-hard optimization objective into a convex upper surrogate, and optimize it follow the concept of adversarial multi-armed bandits.
    \item Design a novel update strategy by combining fair weights computed from loss variance constraints with average weights to ensure server model performance.
    \item Theoretically prove that the server model's generalization performance is upper bounded by the variance of client performance, showing that constraining client performance distribution improves server model generalization. We also provide a convergence analysis for FedMABA, showing it converges to the stationary point at a rate of $\mathcal{O}(\frac{1}{\sqrt{NKT}} + \frac1T)$.
    \item Conduct extensive experiments on Fashion-MNIST, CIFAR-10, and CIFAR-100 within FL, demonstrating that FedMABA maintains competitive server model performance, outperforms baselines in fairness, and exhibits stability across different hyper-parameters and scenarios.
\end{enumerate}

\section{Related Work}
% problems involve balancing exploration (trying different options) and exploitation (choosing the best-known option) when faced with 
% It's widely applied in fields like adaptive clustering for content recommendation \cite{li2016collaborative}, multi-task variance regularization \cite{mao2021banditmtl, yuan2023equitable}, and collaborative computation in federated learning (FL) frameworks \cite{shi2021federated}. 
\textbf{Multi-armed bandit (MAB)} can make decisions when faced with multiple options with unknown results. Our work draws inspiration from the integration of FL and MAB, treating FL clients as entities in a multi-objective optimization problem. Following \cite{mao2021banditmtl, yuan2023equitable}, we introduce MAB to optimize the relaxed NP-hard constraint on client performance disparity, promoting fairness while ensuring the generalization performance of the server model.

\vspace{\baselineskip}

\noindent \textbf{Fair federated learning} is challenging due to data and device heterogeneity \cite{zhou2021towards, wang2023fedeba+}. There are numerous definitions and research directions for fairness in FL, and we focus on fair resource allocation and aggregation (see Def. \ref{Fairness via variance} for details). Fair allocation and aggregation can speed up model convergence, enhance generalization, and motivate client participation. For instance, q-FFL \cite{li2019fair} uses a hyper-parameter $q$ to balance losses among clients, prioritizing those with higher losses. FedMGDA+ \cite{hu2020fedmgda+} seeks aggregation weights that minimize the Euclidean norm of aggregated model updates, while FedFV \cite{wang2021federated} balances gradient update angles.

\vspace{\baselineskip}

\noindent Although there is extensive research on promoting consistency in client performance within FL, most approaches achieve this indirectly without imposing a direct performance constraint across clients, leading to insufficient fairness or compromising server model performance. In this paper, we theoretically prove that the server model's generalization error is upper bounded by client performance disparities. This shows that explicitly constraining performance disparities improves both the generalization performance of the server model and fairness in FL. We introduce the concept of BanditMTL \cite{mao2021banditmtl} into FL, relaxing NP-hard explicit constraints into a convex upper surrogate, and design an FL algorithm to solve it. We also provide a convergence analysis for our proposed algorithm FedMABA, showing it converges to the stationary point at a rate of $\mathcal{O}(\frac{1}{\sqrt{NKT}} + \frac1T)$.

\section{Preliminaries and Metrics}

\begingroup
\setlength{\parskip}{3.pt plus0pt minus1.75pt}
% In this section, we introduce the basic FL paradigm and the definition of FL fairness as analysed in our paper, and then present our performance disparity regularization dedicated to fair FL.
In this section, we introduce the FL setting considered in this work and the basic FL paradigm in Sec. \ref{basic federated learning}. We provide the definition of FL fairness as analysed in our paper in Sec. \ref{fairness definition}, and then present our performance disparity regularization dedicated to fair FL in Sec. \ref{optimization objective}.

\subsection{Basic Federated Learning}
\label{basic federated learning}
We consider the horizontal FL scenario, and the typical optimization paradigm can be formulated as follows:
    {\small
    \begin{align}
    \label{basic fl}
    \min_w\left\{F(w)\triangleq\sum_{i=1}^N p_iF_i(w)\right\} ,
\end{align}}
where {\small ${F}_i(w) \triangleq \mathbb{E}_{\xi \sim D_{i}}\left[F_{i}\left(w, \xi\right)\right]$} is the local objective for client $i$ with client's data distribution {\small $D_i$ }, model parameter $w$, and $p_i$ represents the aggregation weight for client $i$, satisfying {\small \(\sum_{i=1}^N p_i = 1\)}. 

\subsection{Fairness via Variance}
    \label{fairness definition}
    In terms of performance distribution consistency,  we provide the definition of FL fairness following \cite{li2019fair}: 
    \begin{definition}[Fairness via Variance]
    \label{Fairness via variance}
    A server model $w_a$ is defined more fair than $w_b$ if the clients' test performance distribution of $w_a$ across $N$ clients is more uniform than that of $w_b$, i.e. {
    \small $\operatorname{var}\left\{F\left(w_a\right)\right\}<\operatorname{var}\left\{F\left(w_b\right)\right\}$}, where {\small $\operatorname{var}\left\{F\left(\cdot\right)\right\} = \frac{1}{N}\sum_{i=1}^{N}\begin{bmatrix}F_{i}(\cdot)-\frac{1}{N}\sum_{i=1}^{N}F_{i}(\cdot)\end{bmatrix}^{2}$} denotes the test variance of client-specific performance.  
    \end{definition}

\subsection{Our Fair FL Optimization Objective}
\label{optimization objective}
% Due to the heterogeneity of data and devices in FL, simply averaging server model updates can lead to significant performance variance, resulting in improper aggregation of feature information and adversely affecting the performance of the server model.
To ensure consistency in client performance, the most intuitive and direct way is to incorporate the variance in client performance as a constraint into the optimization objective as follows:
{\small
\begin{align}
    \label{fair aggregation formulation}
    % \min_w\left\{F(w)\triangleq\sum_{i=1}^N p_kF_k(w)\right\},
    \min_{w}\left\{F(w;\rho)\triangleq\sum_{i=1}^N p_iF_i(w)+\sqrt{\rho_{Var(F)}}\right\} ,
    % f(w;\rho):=\frac{1}{N}\sum_{i=1}^{N}F_i(x)+\sqrt{\rho_{Var(F)}}  ,
\end{align}}
% \end{small}
where $\rho$ is a constraint hyper-parameter that balance the trade-off between mean empirical loss and variance across clients. The parameter $w$ denotes the statistical shared model parameters between all participants we seek to optimize. Please note that the {\small $F_i(w)$} is practically calculated via the empirical risk, {\small $e.g.$ $F_i(w)=\frac{1}{n_i}\sum_{j=1}^{n_{i}}F_i(w, \xi_i^j)$}, where {\small $\xi_i^j=(x_i^j, y_i^j)$} denotes any data sample held by client $i$, and $n_i$ is the total number of data samples.
\begin{remark}[The difficulty of solving problem \ref{fair aggregation formulation}]
       Directly solving optimization problem \ref{fair aggregation formulation} is NP-hard. In Sec. \ref{section maba}, we detail how we leverage the idea of adversarial multi-armed bandit to optimize the relaxed upper surrogate problem.
\end{remark}

\section{FedMABA: MAB Allocation and Aggregation Method}
\label{section maba}
In this section, we provide theoretical analysis, derivations, and practical algorithm design related to our method. More details on derivations, proofs, and analyses are available at \href{https://github.com/Zacks917/Anonymous}{https://github.com/Zacks917/Anonymous}.
% \begin{enumerate}[leftmargin=12pt,nosep]
% \setlength{\itemsep}{3.5pt}
%     % \item Implements \tang{a bit weak} an fair pruning algorithm based on \textit{MAB}. It adapts to the client's own data, explore the highest channel-wise pruning rewards, achieving pruning through sorting, thus ensuring the performance loss of the pruned model is sufficiently low.
%     % \item Introduces an iterative cumulative aggregation pattern for aggregation of updates to empower clients with larger losses to have more influence, while maintain the expected performance of clients with high performance, thereby ensuring fair treatment of clients with poorer performance in average aggregation.\wang{too long, a what algorithm for what purpose by using what main technique}
%     \item In Sec. \ref{generalization proof}, we prove that explicitly constraining the consistency of client performance distribution benefits the generalization capability of the server model, thus not compromising the performance of the server model.
%     \item In Sec. \ref{sec fair allocation and aggregation}, we provide the detailed derivation and a practical algorithm for FedMABA.
%     \item In Sec. \ref{convergence proof}, we analyze the convergence properties of FedMABA.
% \end{enumerate}

% $F_i(w) \triangleq \mathbb{E}_{\xi \sim D_{i}}\left[F_{i}\left(w, \xi\right)\right]$, and practically calculated via the empirical risk, $i.e.$ $F_i(w)=\frac{1}{n_i}\sum_{j=1}^{n_{i}}F_i(w, \xi_i^j)$, where $n_i$ is the number of mini-batches held by client $i$.
\subsection{Higher Generalization performance via Lower Performance Disparities}
\label{generalization proof}
We will demonstrate in this section why explicit constraints on client performance consistency will not compromise the performance of the server model. Recall that the empirical loss of client can be formulated as {\small $F_i(w)=\frac{1}{n_i}\sum_{j=1}^{n_{i}}F_i(w, \xi_i^j)$}, where {\small $\mathcal{D}_i$} is the data distribution held by client $i$. 
% We use $\hat{F_i}$ to denote the empirical loss of client $i$.
\begin{assumption}[Bounded Loss]
    \label{bounded loss}
    W.L.O.G. for client $i$, {\small $\forall \xi_i^m, \xi_i^n$}, there exists a constant $c$, such that {\small $\lvert F_i(w, \xi_i^m) - F_i(w, \xi_i^n)\rvert<c$}.
\end{assumption}
\begin{definition}[Rademacher Complexity]
    Let {\small $\mathcal{F} = F \circ w \triangleq \{\xi \rightarrow F(w, \xi) \mid w \in \mathcal{W}\}$}, for any given dataset of FL {\small $S=(\xi^1, ..., \xi^n)$}, the set of loss mappings is denoted as {\small $\mathcal{F}_{\mid S} \triangleq \{(F(\xi^1), ..., F(\xi^n)) \mid F\in \mathcal{F} \}$}, the Rademacher Complexity is then defined as follows:
    {\small \begin{align}
        \mathcal{R}(\mathcal{F}_{\mid S}) = \frac{1}{N}\mathbb{E}_{\tau \sim \{\pm 1\}^n}\sup_w\sum_{i=1}^N \frac{1}{n_i}\sum_{j=1}^{n_i}\tau F_i(\xi_i^j),
    \end{align}}
    where \(\tau\) has an equal probability of being either 1 or -1.
\end{definition}

\begin{definition}[Representation of Generalization Error]
Given the data distribution {\small \( D \)} and the training data {\small \( S \)}, the representation of generalization error is defined as the supremum of the difference between the expected risk and the empirical risk over the hypothesis space {\small $\mathcal{W}$}. Formally, it is expressed as:
{\small \begin{align}
    Rep_D(S; \mathcal{W}, \mathcal{F}) = \sup_{w \in \mathcal{W}} (F_D(w) - F_S(w)), 
\end{align}}
where {\small $F_{\mathcal{D}}(w)=\mathbb{E}_{\tilde{\xi} \sim\mathcal{D}}F(w, \tilde{\xi})$}, and {\small $F_S(w) = \frac{1}{\lvert S \rvert} \sum_{\xi \in S} F(w, \xi)$}.

\end{definition}
% \begin{lemma}
%     \label{lemma 1}
%     Assume $\forall \xi_m, \xi_n$, $\lvert F_i(w, \xi_m) - F_i(w, \xi_n)\rvert<c$. Define $ \text{Rep}(\mathcal{H}, S) \triangleq \sup_{w \in \mathcal{H}} \left(F_i(w, \xi_m) - \hat{F}_i(w)  \right), $ where $\mathcal{H}$ is the parameter hypothesis set and \( S \) is the sample space, then for $\forall \delta \in \left[0, 1\right]$, with probability of at least $1-\delta$:
%     \begin{align}
%         Rep(\mathcal{H},S)\leq\mathbb{E}_SRep(\mathcal{H},S)+c\sqrt{\frac{2\ln(2/\delta)}{N}}
%     \end{align}
% \end{lemma}

% \begin{lemma}
%     Define
%     \begin{align}
%         R(\mathcal{F}|_{S}) \triangleq \frac{1}{N}\mathbb{E}_{\tau\sim\{\pm1\}^{n_i}}\sup_{w}
%         \sum_{i=1}^N \frac{1}{n_i}\sum_{j=1}^{n_i}\tau_{i, j}F_i(w_i, \xi_j) .
%     \end{align}
%     Combine Lemma \ref{lemma 1}, for $\forall \delta \in \left[0, 1\right]$, with probability of at least $1-2\delta$:
%     \begin{align}
%         \frac{1}{N}\sum_{i=1}^N(F_i(w_i)-\hat{F}_i(w_i)) \leq 2 R(\mathcal{F}|_{S}) + 3c\sqrt{\frac{2\ln(2/\delta)}{N}}
%     \end{align}
% \end{lemma}

\begin{theorem}[Bounded Generalization Error]
    \label{bounded generalization error}
    Consider a FL system with N clients, and parameter hypothesis space {\small $\mathcal{W}$}. If Assumption \ref{bounded loss} holds. Then, for {\small $\forall \delta \in \left[0, 1\right]$} with probability of at least {\small $1-2\delta$}, for {\small $\forall w \in \mathcal{W}$}, the generalization error can be bounded as:
    {\small \begin{align}
        Rep_D(S; \mathcal{W}, \mathcal{F}) \leq 2\mathcal{R}(\mathcal{F}_{\mid S}) + 3c\sqrt{\frac{2\ln \frac{2}{\delta}}{N}}.
    \end{align}}
\end{theorem}

\begin{corollary}[Bounded Generalization Error]
    \label{extended theorem}
     Under the same conditions as Theorem \ref{bounded generalization error}, let { \small S}, {\small D} be the training set and generalization data distribution, {\small $w_S = ERM_w(S) = argmin_{w\in \mathcal{W}}{F(w;S)}$}, {\small $w^* = argmin_{w\in \mathcal{W}}{F(w;D)}$}. Then, for {\small $\forall \delta \in \left[0, 1\right]$} with probability of at least {\small $1-3\delta$}, for {\small $\forall w \in \mathcal{W}$}, the generalization error can be bounded as:
    {\small \begin{align}
        F_{\mathcal{D}}(w_S)-F_{\mathcal{D}}(w^*)\leq 2\sup_{w \in \mathcal{W}} Var(F)+4c\sqrt{\frac{2\ln(2/\delta)}{N}}, 
    \end{align}} 
    where {\small $Var(F) = \sqrt{\frac{1}{N}\sum_{i=1}^{N}(F_{i}(w)-\frac{1}{N}\sum_{i=1}^{N}F_{i}(w))^2}$}.
\end{corollary}
% $F_i(w) \triangleq \mathbb{E}_{\xi_{i} \sim D_{i}}\left[F_{i}\left(w, \xi_{i}\right)\right]$, and practically we calculate the empirical risk, $i.e.$ $F_i(w)=\sum_{j=1}^{n_{i}}l(w,x_{i,j})$, where $n_i$ is the number of data pairs held by client $i$.

        % F_{\mathcal{D}}(w_S)-F_{\mathcal{D}}(w^*)\leq2R(\mathcal{F}_{\mid S})+4c\sqrt{\frac{2\ln(2/\delta)}{N}}
\begin{remark}
    Theorem \ref{extended theorem} states that under the considered FL setting, the generalization performance error of the server model is upper bounded by the consistency of the empirical loss variance. Therefore, explicitly constraining the loss variance will not compromise the generalization performance of the server model while improving fairness.

\end{remark}
\subsection{Fair Allocation and Aggregation}
\label{sec fair allocation and aggregation}

\begin{algorithm}[tb]
    \caption{FedMABA}
    \label{FedMABA}
    \textbf{Input}: server step size {\small $\eta_s$}, client learning rate {\small $\eta_c$}, MAB step size {\small $\eta_b$}, number of communication rounds {\small $T$}, MAB trade-off value {\small $\alpha$}.\\
    \textbf{Initialize}: server model {\small ${w}^0$}, client model {\small ${w}^0$, $p^0 = (\frac{1}{N}, \frac{1}{N}, ..., \frac{1}{N})$}.\\
    \textbf{for {\small $t = 0$} to {\small $T-1$} do}: 
    \begin{algorithmic}[1] %[1] enables line numbers
        \STATE Server selects a subset {\small $N_t$} of clients and send them server model {\small $w^t$}.
        % \IF {Do adaptive pruning}
        
        \FOR{each client ${i} \in N_t$, in parallel}
        \FOR{{\small $k = 0, ..., K-1$}}
        \STATE {\small $w_{i, k+1}^{t} = w_{i, k}^{t} - \eta_c g_{i, k}^{t};$}
        \ENDFOR
        \STATE Let {\small $\Delta_{i}^{t}=w_{i, K}^{t}-w_{i, 0}^{t}=-\eta_{c}\sum_{k=0}^{K-1}g_{i, k}^{t};$}
        \ENDFOR
        % \FOR{each task $t\in T$, in parallel}
        % \IF{there are clients belonging to task $t$ ($i\in S_{r, t}$) participating in current round}
        % \STATE Server aggregates task $t$ update $\Delta_t^r$ by: \\
        %         $\Delta_t^r = \frac{1}{\lvert S_{r, t} \rvert}\sum_{i\in S_{r, t}} \Delta_{ti}^r$
        % \STATE Server update task $t$: $x_t^{r+1} = x_t^r + \eta \Delta_t^r$
        % \ENDIF
        % \ENDFOR
        \STATE Server do
        \STATE  Extract the corresponding weights {\small $p_{N_t}^t$} of length {\small $\lvert N_t \rvert$ from $p^t$}. \\
                Compute $\lambda_t^*$ with binary search according Eq.\eqref{f lambda}.\\
                Update $p_{N_t}^t$ by Eq.\eqref{iterative for p}. \\
                Obtain $p^{t+1}$ by synchronizing the updates of $p_{N_t}^t$ with $p^t$. \\
                Normalize the MAB allocation weights for {\small $N_t$}: \\
                {\small $\tilde{p}_{i}^{t+1} = \frac{p_{i}^{t+1}}{\sum_{i \in N_t} p_{i}^{t+1}};$} 
        %         $p_{i}^{t+1}=\frac{e^{\frac{1}{1+\lambda_t^*}(\log p_{i}^{m}+\eta_{b}F_i(\Tilde{x}_i^{r+1}))}}{\sum_{t=1}^{T}e^{\frac{1}{1+\lambda_t^*}(\log p_{i}^{m}+\eta_{b}F_i(\Tilde{x}_i^{r+1}))}}$ \\
        %         \textbf{Calculates $x$:} \\
        %         \tiny $x_{m}^{r+1} = x^r + \eta_s \frac{1}{N}\left[\alpha \sum_{i=1}^{N} {p_i^0\Delta_i^r} + (1-\alpha)\sum_{i=1}^{N} {p_i^{m+1}\Delta_i^r}\right]$
        % \STATE Choose $x_{\tilde{m}}^r, \tilde{m}\in \lvert M \rvert$ with best validation performance.
        \STATE Server update $w$ by Eq.\eqref{final objective}: \\ 
                    % $w^{t+1} = w^{t} + \alpha\sum_{i \in N_t}\tilde{p}_{i}^{t+1}\Delta_i^t+(1-\alpha)\frac{1}{\lvert N_t \rvert}\sum_{i \in N_t}\Delta_i^t;$
    \end{algorithmic}
    Output: The server model of best performance in evaluation.
\end{algorithm}

As discussed earlier, solving the optimization problem Eq.\eqref{fair aggregation formulation} is challenging due to its non-convex and NP-hard nature. To circumvent this, following \cite{ben2013robust, bertsimas2018robust}, we introduce a surrogate convex upper bound:
{\small \begin{align}
    \label{objective upperbound}
    \sup_{p\in\mathcal{P}_{\rho, N}} \sum_{i=1}^{N} {p_{i}F_{i}(w, \xi_i)}   = \frac{1}{N}\sum_{i=1}^{N}{F_{i}(w, \xi_i)} + \sqrt{\rho_{Var(F)}} + o(N^{-\frac{1}{2}}) \, ,
\end{align}}
where 
{\small $\mathcal{P}_{\rho, N} := \left\{p\in\mathbb{R}^{N}: \sum_{i=1}^{N}{p_i = 1}, \sum_{i=1}^{N}{p_i\log{({N}p_i)}\leq\rho}\right\}$}, and $\rho$ is a constant with respect to the trade-off between mean empirical loss and fairness across clients. With this surrogate, we reformulate the initial objective as follows:
{\small \begin{align}
    \min_w\sup_{p\in\mathcal{P}_{\rho, N}}\sum_{i=1}^{N}{p_{i}F_{i}(w, \xi_i)} \, ,
    \label{objective of FedMABA}
    \end{align}}
when {\small $F_i(\cdot)$} is convex and the hypothesis space {\small $\mathcal{W}$} is compact, the problem can reach a saddle point. In order to reach it, following the BanditMTL \cite{mao2021banditmtl} approach, we translate the problem into solving $p$ first. To construct the dual problem for optimizing $p$, we employ Legendre Function {\small $\Phi_{p}(p)=\sum_{i=1}^{N}p_{i}\log p_{i}$}. Mirror gradient ascent-descent method are introduced to solve the dual space problem, and we derive the iterative formula for $p_{N_t}$ as below:
{\small \begin{align}
    \label{iterative for p}
    \forall i \in N_t, p_i^{t+1}=\frac{e^{\frac1{1+\lambda_t^*}(\log p_i^t+\eta_b F_i(w_i^t, \xi_i))}}{\sum_{i \in N_t} e^{\frac{1}{1+\lambda_t^*}(\log p_i^t+\eta_b F_i(w_i^t, \xi_i))}}\, ,
\end{align}}
where {\small $N_t$} is the set of selected clients for communication round {\small $t$}, {\small $\eta_b$} is the step size for MAB, and {\small $\lambda_t^*$} is the kernel of {\small $f(\lambda)$} formulated as:
{\small \begin{align}
    \label{f lambda}
    f(\lambda)=& \begin{aligned}\frac{\sum_{i \in N_t}(\log q_i^t){q_i^t}^{(\frac{1}{1+\lambda})}}{\sum_{i \in N_t}(1+\lambda){q_i^t}^{(\frac{1}{1+\lambda})}}-\log\sum_{i \in N_t}{q_i^t}^{(\frac{1}{1+\lambda})}\end{aligned} +\log{N}-\rho \,,
\end{align}}
where {\small $q_{i}^t=e^{(\log p_{i}^{t}+\eta_{b}F_i(w_i^t, \xi_i))}$}. Following \cite{mao2021banditmtl}, we use binary search algorithm to find the kernel {\small $\lambda_t^*$}. 

After obtaining the partial MAB weights {\small $p_{N_t}^{t+1}$}, synchronize it to $p^{t+1}$ first. Then we use the hyper-parameter {\small $\alpha$} to combine the normalized MAB weights {\small $\tilde{p}_{N_t}^{t+1} = \frac{p_{i}^{t+1}}{\sum_{i \in N_t} p_{i}^{t+1}}$} with the average aggregation to obtain the final update formula as follows:
{\small \begin{align}
    \label{final objective}
    w^{t+1} = w^{t} + \alpha\sum_{i \in N_t}\tilde{p}_{i}^{t+1}\Delta_i^t+(1-\alpha)\frac{1}{\lvert N_t \rvert}\sum_{i \in N_t}\Delta_i^t
    \, ,
\end{align}}
where {\small $\tilde{p}_{i}^{t+1}$} is the $i$-th normalized MAB weight of clients set $N_t$. It is worth noting that {\small $\alpha$} is to make a trade-off between average aggregation and MAB aggregation. This is because in practical FL training, it is hard to ensure that all clients participating in each communication round, and not all weights in {\small $p^t$} will be updated in one communication round. Therefore, we introduce average aggregation to mitigate the possible bias caused by partial updates of MAB weights.

\subsection{Convergence Analysis of FedMABA}
\label{convergence proof}
To provide the convergence analysis for FedMABA, following \cite{wang2023delta, wang2020tackling}, we make the assumptions below: (1) Each objective function of clients is Lipschitz smooth with constant $L$. (2) The stochastic gradient calculated by each client can be an unbiased estimator of the clients' gradient, and the expectation of error can be bounded by $\sigma^2$. (3) The dissimilarity of clients' gradient can be bounded with constants $\gamma$ and $A$. (4) The divergence between FedMABA aggregation weights and FedAvg ones can be bounded by a constant $\kappa$. 
\begin{theorem}[Convergence Analysis of FedMABA]
    Under Assumptions above, the total communication rounds T is pre-determined. Let $\eta_s$, $\eta_c$ be the server step size and client step size. If $\eta_c$ is small enough such that {\small $\eta_c < \min \left(\frac{1}{8LK}, C\right)$}, where {\small $\frac12-20L^2\frac1N\sum_{i=1}^NK^2\eta_L^2(\gamma^2+1)(\kappa \gamma^2+1)>C>0$} and {\small $\eta \leq \frac{1}{\eta_sL}$}, then with the proper setting {\small $\eta_{c}=\mathcal{O}\left(\frac{1}{\sqrt{T}KL}\right)$} and {\small $\eta_s=\mathcal{O}\left(\sqrt{NK}\right)$} the convergence rate can be formulated as follows:
    {\small \begin{align}
        % \min_{t\in[T]}\mathbb{E}\|\nabla F(\boldsymbol{w}^{(t)})\|^{2}\leq\mathcal{O}\left(\frac{F^0-F^*}{\sqrt{NKT}}\right)+\mathcal{O}\left(\frac{4\sqrt{N}\sigma^{2}L}{\sqrt{KT}}\right)+ \nonumber \\
        % \mathcal{O}\left(\frac{6NL^2(K-1)\sigma^{2}}{KT}\right)+\mathcal{O}\left(\frac{12NL^2(K-1)A^2}{T}\right) .
        \min_{t\in[T]}\mathbb{E}\left\|\nabla f\left(\boldsymbol{w}_t\right)\right\|^2\leq\mathcal{O}\left(\frac{1}{\sqrt{NKT}}+\frac{1}{T}\right).
    \end{align}}
\end{theorem}

\section{Experiments}
This section details our experimental setup and highlights the superior fairness of FedMABA over other baseline methods.

\subsection{Baselines, Datasets, Models, and Hyper-parameters}  To evaluate the performance of FedMABA, our baseline models include the fundamental FL algorithm FedAvg \cite{mcmahan2017communication}, as well as classic fairness algorithms AFL \cite{mohri2019agnostic}, q-FFL \cite{li2019fair}, TERM \cite{li2020tilted}, FedMGDA+ \cite{hu2020fedmgda+}, FedProx \cite{li2020federated} and PropFair \cite{zhang2022proportional}. The hyper-parameters are detailed in Table~\ref{hyper table}. We adapt all baseline algorithms to evaluate the test performance denoted as Global Acc., performance disparities denoted as Var.(Fairness), and the average worst $5\%$ and best $5\%$ of client performance. All baselines are evaluated based on publicly available datasets: FASHION-MNIST \cite{xiao2017fashion}, CIFAR-10, and CIFAR-100 \cite{krizhevsky2009learning}. We devided the datasets into Non-IID (Non-Independently and Identically Distributed) forms using two different methods: (1) Following the setting of \cite{wang2021federated}, every clients obtain two shards of differently labeled data. (2) Following Latent Dirichlet Allocation (LDA) with $\alpha=0.5$. All clients perform local updates based on Stochastic Gradient Descent (SGD) with learning rate $\eta = 0.1$ of $0.999$ decay per communication round. The Fashion-MNIST experiments are running over 1000 rounds, while 2000 on CIFAR-10 and CIFAR-100. A single client update consists of one epoch (stochastic steps $K=10$) with local batch size $B = 50$. The reported results are averaged over 3 runs with different random seeds. We highlight the best and the second-best results by using \textbf{bold font} and \textcolor{blue}{blue text}.

\begin{table}[!t]
    \small
    \caption{\small hyper-parameters of all experimental algorithms.}
    \resizebox{.4\textwidth}{!}{
    \begin{tabular}{ll}
    \toprule
        Algorithm & Hyper Parameters \\ \midrule
        $q-FFL$ &  $q \in \{0.005, 0.01, 0.1, 0.5, 5, 15\}$ \\ 
        $PropFair$ & $M \in \{0.2, 5.0\}, \epsilon = 0.2$ \\
        $AFL$ & $\lambda \in \{0.01, 0.1, 0.5, 0.7\}$ \\ 
        $TERM$ & $T \in \{0.1, 0.5, 0.8\}$ \\ 
        $FedMGDA+$ & $\epsilon \in \{0.1, 1.0\}, \eta=1.0$ \\
        $FedProx$ & $\mu = 0.1 $ \\
        % $Scaffold$ & $\eta = 1.0 $,${\mu = 1.0}$ \\
        \multirow{2}{*}{$FedMABA$} & $\eta_b \in \{1.0, 0.7, 0.5, 0.3, 0.1\}$, \\
        & $\alpha \in \{0.1, 0.3, 0.5, 0.7, 0.9\}, \rho=1.0$ \\ \bottomrule
        % FedEBA & $ T \in \{1.0, 0.5, 0.1, 0.05, 0.01\}$ \\ 
    \end{tabular}
    }
    \centering
    \label{hyper table}
    \vspace{-1.em}
\end{table}

\begin{table*}[!t]
 \small
 \centering
 \caption{\small Performance of algorithms on Fashion-MNIST and CIFAR-10.} 

 \fontsize{7.6}{11}\selectfont 
 % \vspace{-2.em}
 
 \resizebox{1.\textwidth}{!}{
  \begin{tabular}{l l l l l l l l l c c}
   \toprule
   \multirow{2}{*}{Algorithm} & \multicolumn{4}{c}{Fashion-MNIST (MLP)} & \multicolumn{4}{c}{CIFAR-10 (CNN)}\\
   \cmidrule(lr){2-5} \cmidrule(lr){6-9} 
                & Fairness $\downarrow$  & Global Acc.  $\uparrow$        & Worst 5\% $\uparrow$ & Best 5\% $\uparrow$ & Fairness $\downarrow$ & Global Acc.  $\uparrow$    & Worst 5\% $\uparrow$ & Best 5\% $\uparrow$ \\
   \midrule
FedAvg                        & 50.14{\transparent{0.5}±3.91} & 85.66{\transparent{0.5}±0.13}   & 70.30{\transparent{0.5}±0.70} & 96.70{\transparent{0.5}±0.10} & 72.12{\transparent{0.5}±2.30} & 68.41{\transparent{0.5}±0.12} & 50.20{\transparent{0.5}±0.49} & 83.67{\transparent{0.5}±0.34} \\ 
{FedSGD}                       & 47.89{\transparent{0.5}±3.37}  & 85.42{\transparent{0.5}±0.35}   & 69.80{\transparent{0.5}±0.40} & 96.20{\transparent{0.5}±0.12} & 70.66{\transparent{0.5}±3.15} & 67.95{\transparent{0.5}±0.30}  & 48.78{\transparent{0.5}±0.90} & 82.85{\transparent{0.5}±0.32} \\ 
% \midrule
q-FFL$|_{q=0.005}$      & 47.65{\transparent{0.5}±0.95}   & 85.52{\transparent{0.5}±0.29}  & 69.80{\transparent{0.5}±0.60} & 97.40{\transparent{0.5}±0.36} & 71.19{\transparent{0.5}±0.64} & \textcolor{blue}{68.62{\transparent{0.5}±0.18}}  & 50.40{\transparent{0.5}±0.40} & 82.25{\transparent{0.5}±0.10}\\
% q-FFL$|_{q=0.01}$             & 86.62{\transparent{0.5}± 0.03} & 58.11{\transparent{0.5}± 3.21}  & 71.36{\transparent{0.5}± 1.98} & 95.29{\transparent{0.5}±0.27 } & 68.85{\transparent{0.5}± 0.03} & 95.17{\transparent{0.5}± 1.85}  & 48.20{\transparent{0.5}±0.80} & 84.10{\transparent{0.5}±0.10 }\\
% \midrule
FedMGDA+$|_{\eta=1.0, \epsilon=0.1}$   & 52.37{\transparent{0.5}±1.79}	 & 85.67{\transparent{0.5}±0.60}  & 69.60{\transparent{0.5}±0.37}	 & \textbf{99.00{\transparent{0.5}±0.23}} & 77.27{\transparent{0.5}±1.06} & 68.25{\transparent{0.5}±0.13}& 48.93{\transparent{0.5}±1.01}& \textcolor{blue}{84.07{\transparent{0.5}±0.34}}\\
FedMGDA+$|_{\eta=0.1, \epsilon=0.1}$   & 76.61{\transparent{0.5}±0.24}  & 85.52{\transparent{0.5}±0.08}& 64.80{\transparent{0.5}±0.12} & \textcolor{blue}{98.40{\transparent{0.5}±0.20}}	& 69.69{\transparent{0.5}±1.30} & \textbf{68.73{\transparent{0.5}±0.48}} & 51.00{\transparent{0.5}±0.23} & \textbf{85.40{\transparent{0.5}±0.56}} \\
% \midrule
% $Ditto|_{\lambda_t^*=0.0} $        & 86.37{\transparent{0.5}±0.13 & 55.56{\transparent{0.5}±5.43  & 69.20{\transparent{0.5}±0.37 & 95.79{\transparent{0.5}±0.38 & 60.11{\transparent{0.5}±4.41 & 85.99{\transparent{0.5}±7.13  & 42.20{\transparent{0.5}±2.20 & 77.90{\transparent{0.5}±4.90 \\
% $Ditto|_{\lambda_t^*=0.5} $        & 86.29{\transparent{0.5}±0.25 & 56.57{\transparent{0.5}±2.11  & 69.05{\transparent{0.5}±0.76 & 95.85{\transparent{0.5}±0.36 & 62.65{\transparent{0.5}±1.96 & 104.10{\transparent{0.5}±10.39 & 42.67{\transparent{0.5}±3.45 & 81.27{\transparent{0.5}±0.82 \\ 
% \midrule
TERM$|_{T=0.1}   $         & 317.82{\transparent{0.5}±4.49}  & 69.41{\transparent{0.5}±0.92}& 25.60{\transparent{0.5}±1.56} & 92.40{\transparent{0.5}±0.40} & 199.30{\transparent{0.5}±4.42} & 30.77{\transparent{0.5}±0.15} & 1.80{\transparent{0.5}±0.10} & 59.00{\transparent{0.5}±0.37}   \\ 
AFL$|_{\lambda=0.05} $        & 49.97{\transparent{0.5}±1.02}  & 85.63{\transparent{0.5}±0.09}  & 69.20{\transparent{0.5}±0.53} & 96.60{\transparent{0.5}±1.20} & 64.54{\transparent{0.5}±1.55} & 66.39{\transparent{0.5}±0.17}  & 50.80{\transparent{0.5}±0.43} & 82.08{\transparent{0.5}±0.16}\\
AFL$|_{\lambda=0.5} $         & 56.69{\transparent{0.5}±1.25}  & 85.79{\transparent{0.5}±0.34}  & 67.80{\transparent{0.5}±0.94}& 98.33{\transparent{0.5}±0.27} & 68.92{\transparent{0.5}±0.78} & 68.12{\transparent{0.5}±0.44}  & 51.80{\transparent{0.5}±0.26} & 83.00{\transparent{0.5}±0.20 }\\ 
% \midrule
PropFair$|_{M=5.0, thres=0.2}$ & 89.72{\transparent{0.5}±4.68} & 83.42{\transparent{0.5}±0.45}  & 58.60{\transparent{0.5}±1.22} & 95.80{\transparent{0.5}±0.40}  & 79.64{\transparent{0.5}±2.44} & 65.82{\transparent{0.5}±0.39}  & 43.80{\transparent{0.5}±2.15}& 81.00{\transparent{0.5}±0.65}\\ 
% \midrule
FedProx$|_{\mu=0.1}   $        & 49.89{\transparent{0.5}±2.77}   & 85.15{\transparent{0.5}±0.54} & 70.40{\transparent{0.5}±0.59} & 97.20{\transparent{0.5}±0.38}   & 66.05{\transparent{0.5}±1.67} & 68.29{\transparent{0.5}±0.38}& 51.94{\transparent{0.5}±0.38} & 83.20{\transparent{0.5}±0.86} \\ 
% \midrule

\midrule

% FedMABA$|_{\epsilon=0.5, \eta_b=0.5, \p_r=0.2}$ & \textbf{87.14{\transparent{0.5}±0.19}} & \textcolor{blue}{52.29{\transparent{0.5}±4.34}}  &{70.20{\transparent{0.5}±1.47}} & 97.80{\transparent{0.5}±0.19} & 68.27{\transparent{0.5}±0.25} & \textbf{58.69{\transparent{0.5}±4.39}}  & \textbf{52.40{\transparent{0.5}±1.28} }& 83.72{\transparent{0.5}±0.52} \\ 
FedMABA$|_{\alpha=0.5, \eta_b =0.5}$ & \textcolor{blue}{34.54{\transparent{0.5}±1.12}} & \textbf{86.02{\transparent{0.5}±0.06}}  &\textcolor{blue}{71.98{\transparent{0.5}±0.40}} & 95.88{\transparent{0.5}±0.33} & \textcolor{blue}{62.52{\transparent{0.5}±0.74}} & 68.58{\transparent{0.5}±0.24}  & \textcolor{blue}{52.08{\transparent{0.5}±0.61} }& 82.74{\transparent{0.5}±0.40} \\
FedMABA$|_{\alpha=0.8, \eta_b =0.5}$ & \textbf{31.43{\transparent{0.5}±0.97}}  & \textcolor{blue}{85.88{\transparent{0.5}±0.20}}  &\textbf{72.60{\transparent{0.5}±0.36}} & 95.60{\transparent{0.5}±0.41} & \textbf{60.84{\transparent{0.5}±1.89}}  & 68.43{\transparent{0.5}±0.33}  & \textbf{52.27{\transparent{0.5}±0.62} }& 82.47{\transparent{0.5}±0.98} \\ 
% FedMABA$|_{\epsilon=0.1, \eta_b=0.1}$ & \textbf{87.14{\transparent{0.5}±0.19}} & \textcolor{blue}{52.29{\transparent{0.5}±4.34}}  &{70.20{\transparent{0.5}±1.47}} & 97.80{\transparent{0.5}±0.19} & 68.27{\transparent{0.5}±0.25} & \textbf{58.69{\transparent{0.5}±4.39}}  & \textbf{52.40{\transparent{0.5}±1.28} }& 83.72{\transparent{0.5}±0.52} \\
\bottomrule
\end{tabular}}
\label{Performance table}
\end{table*}

\subsection{Numerical Results}

\begin{enumerate}[label=(\alph*)]
    \item \textbf{Performance Comparison of Algorithms.} We evaluated algorithms using a two-layer MLP on Fashion-MNIST and a two-layer CNN on CIFAR-10 and CIFAR-100, comparing both partial and full client participation scenarios. As shown in Table~\ref{Performance table}, Table~\ref{cifar100table}, and Figure~\ref{cifar100 figure}, FedMABA outperforms other baselines in fairness while maintaining global accuracy. Specifically:
    \begin{itemize}
    \item \textbf{\emph{FedMABA's variance is significantly smaller than other algorithms.}} FedMABA achieves at least \textbf{33\%} better variance on Fashion-MNIST, \textbf{6\%} on CIFAR-10, and even \textbf{1.5x} compared to the second-best baseline. On CIFAR-100, FedMABA shows \textbf{8\%} better variance, indicating greater fairness. Experiments on Fashion-MNIST (Fig. \ref{full fashion}) and CIFAR-10 (Fig. \ref{par cifar}) demonstrate FedMABA’s \textbf{better} and more \textbf{stable} client performance consistency, regardless of participation levels.
    \item \textbf{\emph{FedMABA doesn't compromise server model performance.}} Despite significantly improving client performance consistency, FedMABA’s server model generalization performance remains above average.
    \end{itemize}
    \item \textbf{Impact of Hyper-parameters.} To assess hyper-parameters’ impact on FedMABA, we varied $\alpha$ and $\eta_b$ in Fashion-MNIST experiments. Fig. \ref{ablation study Fairness} shows FedMABA’s stability under hyper-parameter tuning. A larger $\eta_b$ allocates more resources to weaker clients, and prioritizing MAB weights generally improves fairness.
\end{enumerate}

\begin{figure}[!t]
	\centering
	\begin{minipage}[b]{0.48\textwidth}
		\centering
		% 第一张图片及其副标题
		\subcaptionbox{Full participation on Fashion-MNIST\label{full fashion}}{%
			\includegraphics[width=.475\textwidth]{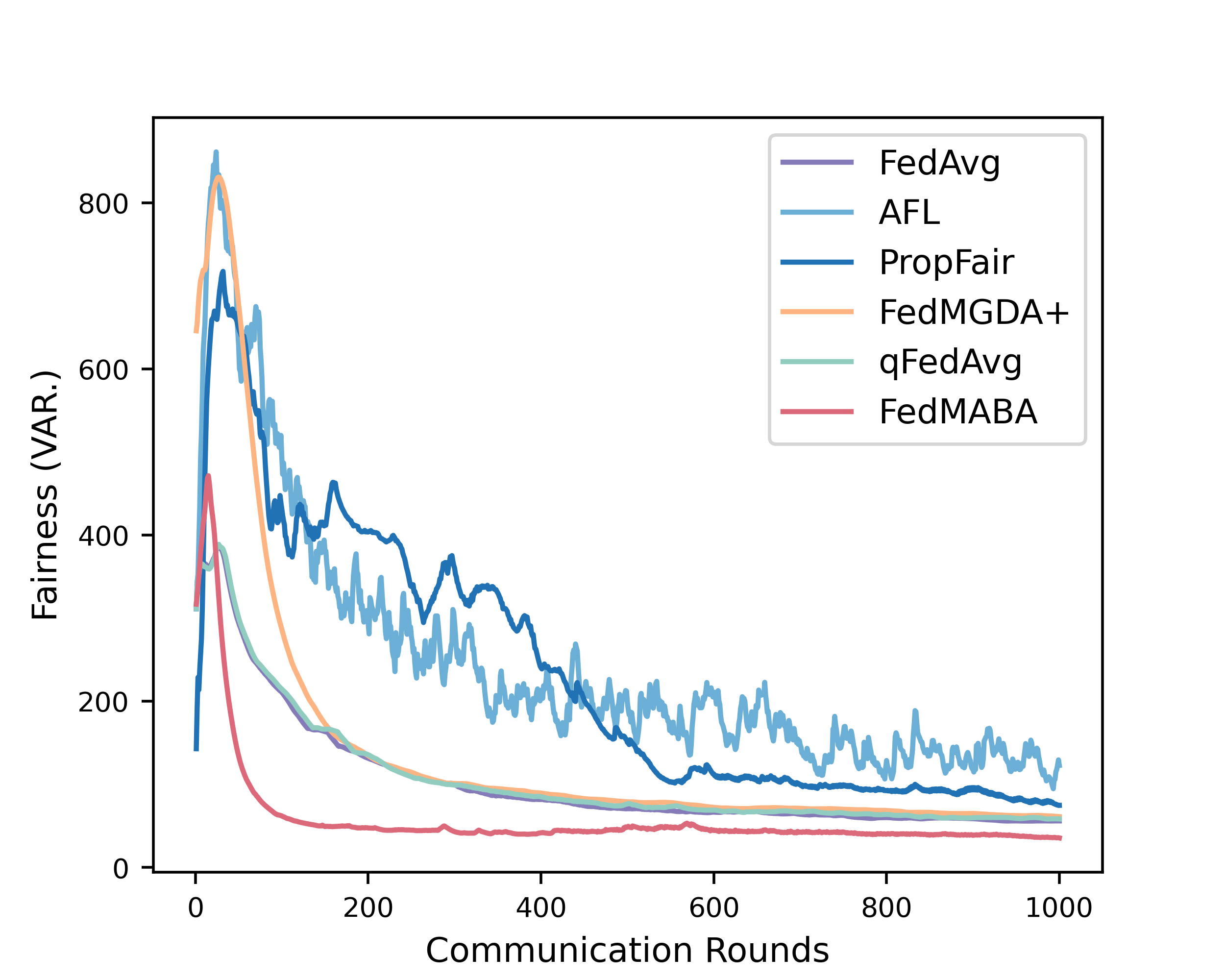}
		}
		\subcaptionbox{Partial participation on CIFAR-10\label{par cifar}}{%
			\includegraphics[width=.475\textwidth]{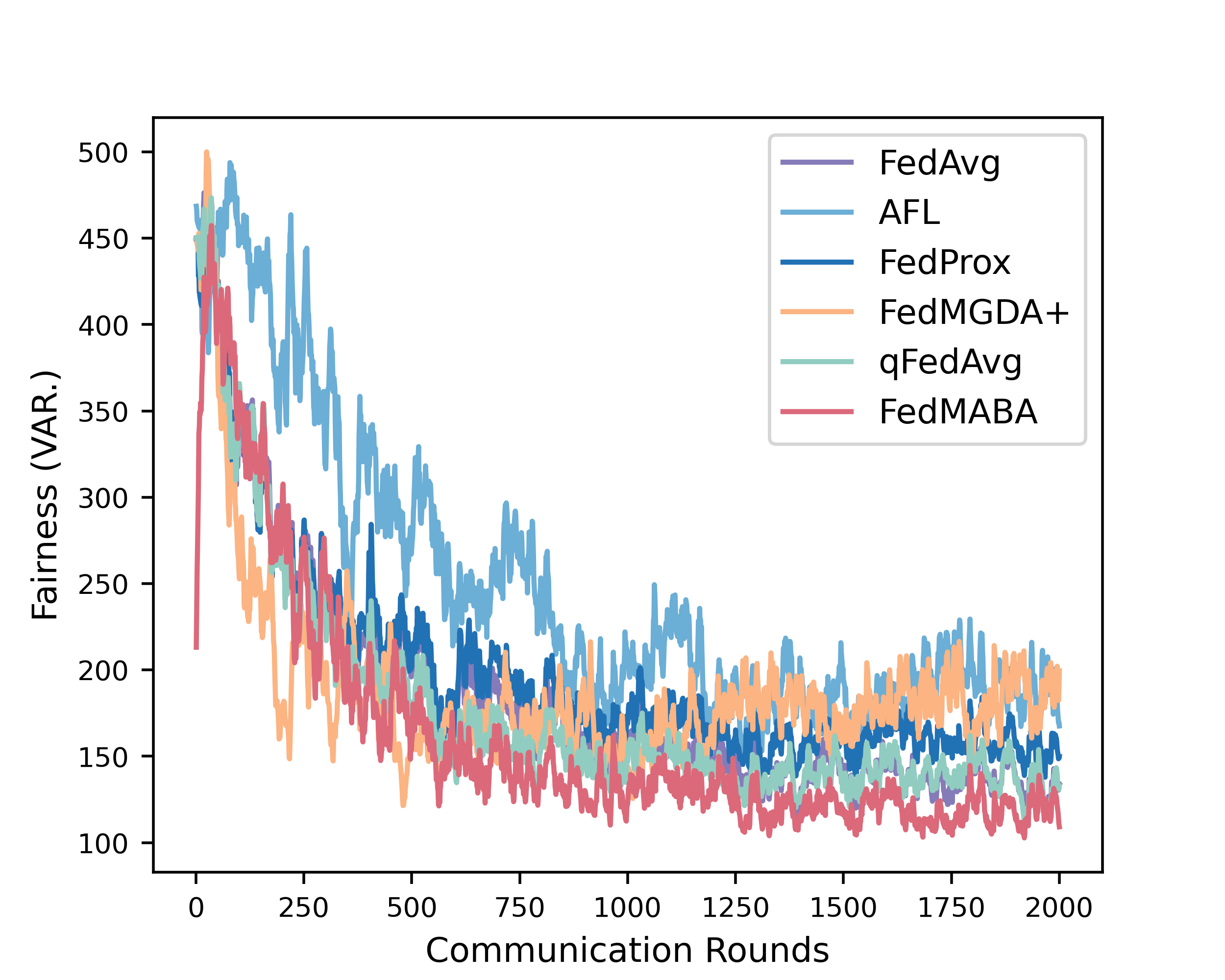}
		}
	\end{minipage}
	\caption{Fairness Performance of Algorithms.} % 简短标题用于列表
	\label{cifar100 figure}
\end{figure}

% \begin{figure}[!t]
% 	\centering
% 	\begin{minipage}[b]{0.48\textwidth}
% 		\centering
% 		\subcaptionbox{Fairness performance for different $\eta_b$ values\label{fairness ab eta}}{% 
% 			\includegraphics[width=.48\textwidth]{current_exp_figure/cifar10/abofeta.png}
% 		}
% 		\subcaptionbox{Fairness performance for different $\alpha$ values\label{fairness ab alpha}}{%
% 			\includegraphics[width=.48\textwidth]{current_exp_figure/cifar10/abofalpha.png}
% 		}
% 	\end{minipage}
% 	\caption{\textbf{Performance of variance on FedMABA.} Ablation Study of $\eta_b$ and $\alpha$. Left: Fixing $\alpha = 0.1$ and examining fairness performance for different $\eta_b$ values. Right: Fixing $\eta_b = 0.5$ and evaluating fairness performance for different $\alpha$ values.}
% 	\label{ablation study Fairness}
% \end{figure}

\begin{figure}[!t]
	\centering
	\begin{minipage}[b]{0.48\textwidth}
		{
			\includegraphics[width=.48\textwidth,]{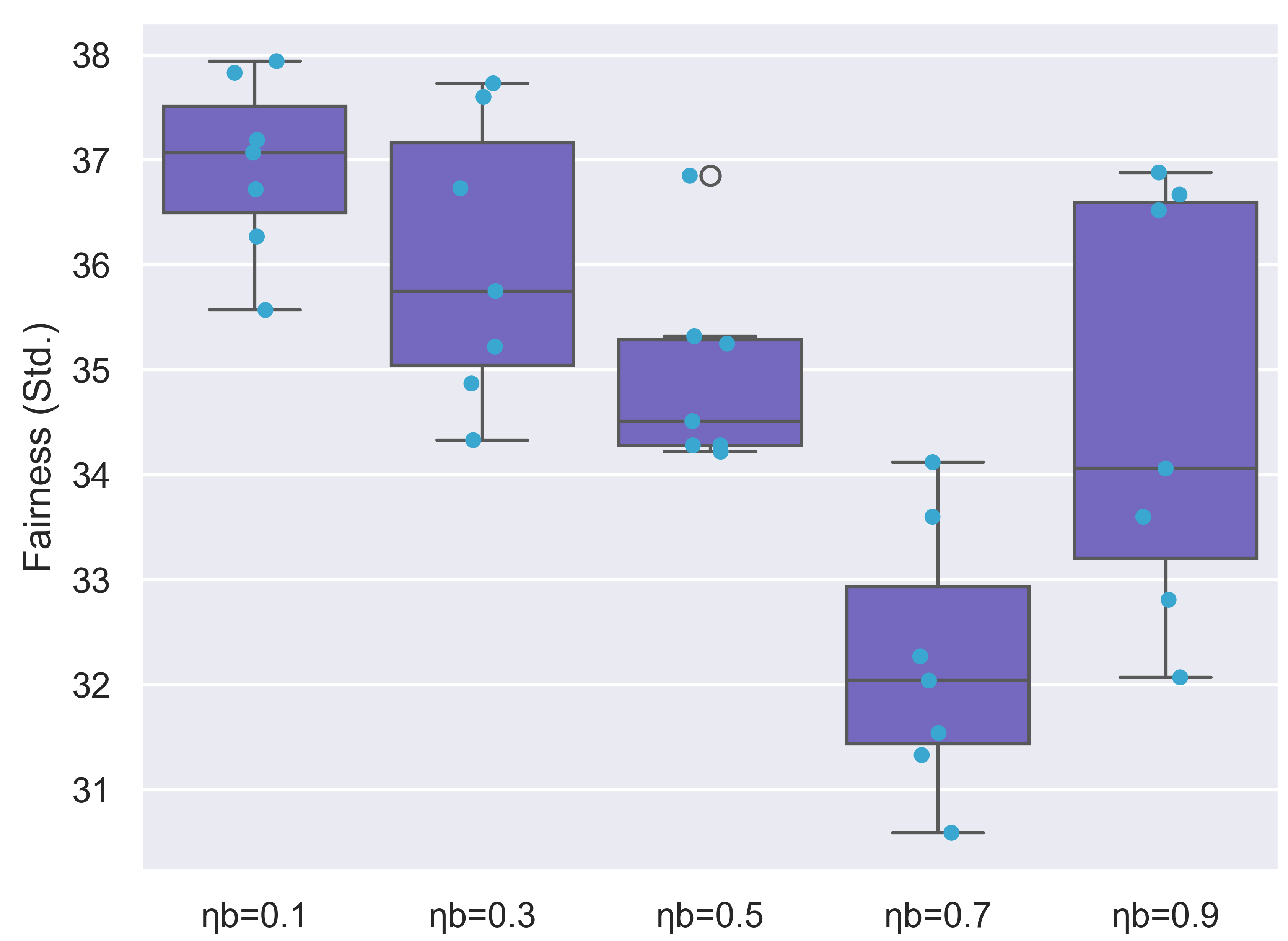}
			\includegraphics[width=.48\textwidth,]{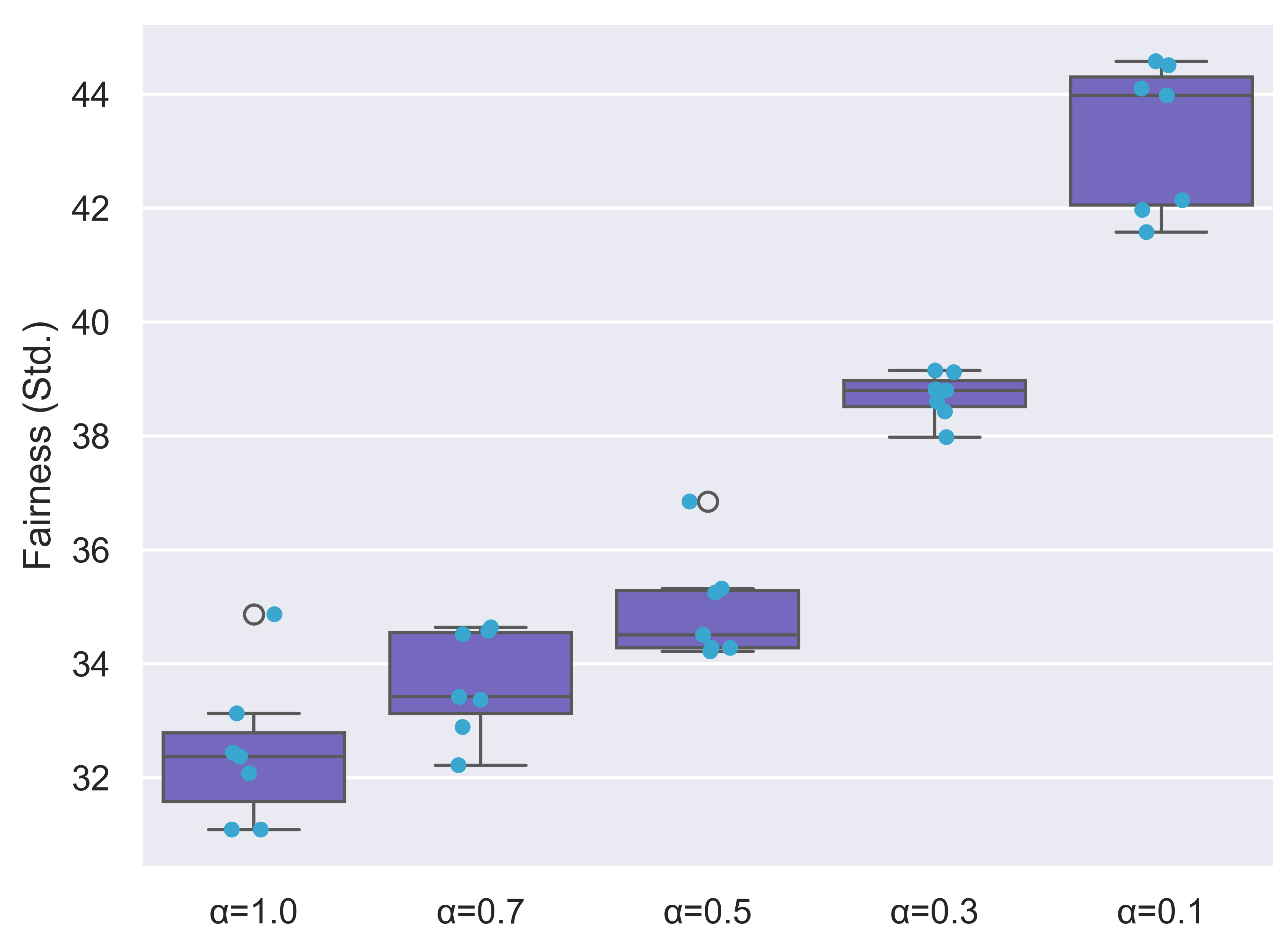}
			\label{fairness ab on cifar10}
	}
	\end{minipage}
     \caption{\textbf{Performance of variance on FedMABA.} Ablation Study of $\eta_b$ and $\alpha$. Left: Fixing $\alpha = 0.5$ and examining fairness performance for different $\eta_b$ values. Right: Fixing $\eta_b = 0.5$ and evaluating fairness performance for different $\alpha$ values.}
	\label{ablation study Fairness}
\end{figure}
\looseness=-1

\begin{table}[!t]
\small
\centering
\caption{Performance of algorithms on CIFAR-100.} 
 % \fontsize{7.6}{11}\selectfont
\label{cifar100table}
\resizebox{.45\textwidth}{!}{
\begin{tabular}{l c c c c}
\toprule
\multirow{2}{*}{\centering Algorithm} & \multicolumn{4}{c}{CIFAR-100 (CNN)}     \\ \cmidrule{2-5}
                 & Fairness $\downarrow$ & Global Acc. $\uparrow$ & Worst 5\% $\uparrow$ & Best 5\%  $\uparrow$  \\
                 \midrule
FedAvg & 25.49{\transparent{0.5}±0.26} & \textbf{56.95{\transparent{0.5}±0.12}} & 45.16{\transparent{0.5}±0.51} & \textcolor{blue}{67.59{\transparent{0.5}±0.23}} \\ 
q-FedAvg & 24.58{\transparent{0.5}±0.31} & 56.49{\transparent{0.5}±0.04} & 44.69{\transparent{0.5}±0.15} & 67.42{\transparent{0.5}±0.47} \\ 
FedProx & 23.47{\transparent{0.5}±0.55} & \textcolor{blue}{56.56{\transparent{0.5}±0.24}} & \textcolor{blue}{46.05{\transparent{0.5}±0.06}} & 65.97{\transparent{0.5}±0.35} \\ 
FedMGDA+ & 23.72{\transparent{0.5}±0.27} & 56.08{\transparent{0.5}±0.11} & 45.60{\transparent{0.5}±0.41} & 66.54{\transparent{0.5}±0.70} \\ 
AFL & \textcolor{blue}{22.73{\transparent{0.5}±0.14}} & 55.59{\transparent{0.5}±0.17} & 45.56{\transparent{0.5}±0.34} & 65.23{\transparent{0.5}±0.15} \\ 
\midrule
FedMABA & \textbf{22.27{\transparent{0.5}±0.08}} & 56.47{\transparent{0.5}±0.29} & \textbf{47.29{\transparent{0.5}±0.10}} & \textbf{67.61{\transparent{0.5}±0.44}} \\ 
\bottomrule
\end{tabular}
}
\vspace{-1.em}
\end{table}

\section{Conclusion and Future Works}

% We propose FedMABA, the first framework that combines Multi-Armed Bandit (MAB) and Federated Learning (FL) to address fairness issues in FL. By designing a novel aggregation approach, FedMABA overcomes the limitations of traditional MAB-based multi-task learning algorithms in the FL scenario. This ensures improved fairness while maintaining model convergence efficiency. Experimental results demonstrate the stability of FedMABA across hyper-parameters, its outstanding performance in exploring the globally optimal fair direction, and its effective pruning capabilities. Due to the nature of the search, iteration, and validation in MAB methods, exploring ways to achieve better results in shorter rounds and thus further promote convergence efficiency represents a valuable direction. 
Through MAB weights Allocation and aggregation approach, FedMABA enhances fairness while maintaining the convergence efficiency and generalization performance of server model. Experimental results showcase FedMABA's superiority in terms of fairness. However, malicious clients may falsely report or manipulate model losses to gain more attention under such scenario. Ensuring the robustness of FedMABA in the face of adversarial attacks in FL is a direction for future work.

\section*{Acknowledgment}
This work is supported in part by funding from the Shenzhen Institute of Artificial Intelligence and Robotics for Society, in part by the Shenzhen Key Lab of Crowd Intelligence Empowered Low-Carbon Energy Network (Grant No. ZDSYS20220606100601002), in part by the Shenzhen Stability Science Program 2023, and in part by the Guangdong Provincial Key Laboratory of Future Networks of Intelligence (Grant No. 2022B1212010001). \textit{(Corresponding author: Xiaoying Tang.)}

\bibliographystyle{IEEEtran}
% \bibliography{IEEEabrv,main}
\bibliography{main}

% \newpage
% \onecolumn
% \input{proof}

% \begin{thebibliography}{00}
% \bibitem{b1} G. Eason, B. Noble, and I. N. Sneddon, ``On certain integrals of Lipschitz-Hankel type involving products of Bessel functions,'' Phil. Trans. Roy. Soc. London, vol. A247, pp. 529--551, April 1955.
% \bibitem{b2} J. Clerk Maxwell, A Treatise on Electricity and Magnetism, 3rd ed., vol. 2. Oxford: Clarendon, 1892, pp.68--73.
% \bibitem{b3} I. S. Jacobs and C. P. Bean, ``Fine particles, thin films and exchange anisotropy,'' in Magnetism, vol. III, G. T. Rado and H. Suhl, Eds. New York: Academic, 1963, pp. 271--350.
% \bibitem{b4} K. Elissa, ``Title of paper if known,'' unpublished.
% \bibitem{b5} R. Nicole, ``Title of paper with only first word capitalized,'' J. Name Stand. Abbrev., in press.
% \bibitem{b6} Y. Yorozu, M. Hirano, K. Oka, and Y. Tagawa, ``Electron spectroscopy studies on magneto-optical media and plastic substrate interface,'' IEEE Transl. J. Magn. Japan, vol. 2, pp. 740--741, August 1987 [Digests 9th Annual Conf. Magnetics Japan, p. 301, 1982].
% \bibitem{b7} M. Young, The Technical Writer's Handbook. Mill Valley, CA: University Science, 1989.
% \end{thebibliography}
% \vspace{12pt}
% \color{red}
% IEEE conference templates contain guidance text for composing and formatting conference papers. Please ensure that all template text is removed from your conference paper prior to submission to the conference. Failure to remove the template text from your paper may result in your paper not being published.

\end{document}